\documentclass[twoside,11pt]{article}
\usepackage{jair, rawfonts}
\usepackage{amsmath}
\usepackage{booktabs}
\usepackage{adjustbox}
\usepackage{geometry}
\usepackage{multirow}
\usepackage{url}
\usepackage{graphicx}
\usepackage{soul}
\usepackage{xspace}
\usepackage[normalem]{ulem}
\usepackage{threeparttable}
\usepackage{enumitem}
\usepackage{makecell}
\usepackage{subfigure}
\usepackage{subcaption}
\usepackage[title]{appendix}
\usepackage{tikz}
\usepackage{hyperref}
\usepackage{xcolor}
\usepackage[utf8]{inputenc}
\usepackage{url}

\usepackage[authoryear]{natbib}

\usepackage[utf8]{inputenc}
\usepackage{newunicodechar}
\newunicodechar{≥}{\geq}

\newcommand*\circled[1]{\tikz[baseline=(char.base)]{
    \node[shape=circle,fill=black,inner sep=0.3pt] (char) {\textcolor{white}{#1}};}}

\usepackage{hyperref}
\hypersetup{
    colorlinks=true,
    linkcolor=blue,
    citecolor=blue,
    filecolor=blue,
    urlcolor=blue
}
\usepackage{listings}
\usepackage{amsfonts}

\lstdefinestyle{mystyle}{
    backgroundcolor=\color{white},    % Background color
    commentstyle=\color{black},       % Comment style
    keywordstyle=\color{teal},        % Keywords in blue
    numberstyle=\tiny\color{gray},    % Line numbers style (now removed)
    stringstyle=\color{darkred},      % String literals in dark red
    basicstyle=\ttfamily\footnotesize,% Main font for code
    breaklines=true,                  % Line breaking enabled
    numbers=none,                     % No line numbers
    frame=single,                     % Frame around each code block
    showspaces=false,                 % Don't show spaces explicitly
    showstringspaces=false,           % Don't show spaces in strings
    showtabs=false,                   % Don't show tabs explicitly
    tabsize=2,                        % Tab size is 2 spaces
    morekeywords= {Task , Decomposition, KPM, Scores}
}

\ShortHeadings{RDoLT Reasoning}
{Kaleem Ullah Qasim, Jiashu Zhang, Tariq Alsahfi, Ateeq Ur Rehman Butt}
\firstpageno{1}

\begin{document}

\title{Recursive Decomposition of Logical Thoughts: Framework for Superior Reasoning and Knowledge Propagation in Large Language Models}

\author{\name Kaleem Ullah Qasim \email kaleem@my.swjtu.edu.cn \\
       \addr Southwest Jiaotong University\\
       % % % Chengdu Sichuan, 611756 China
       \AND
       \name Zhang Jiashu \email jszhang@home.swjtu.edu.cn \\
       \addr Southwest Jiaotong University \\
       \AND
       \name Tariq Alsahfi \email tmalsahfi@uj.edu.sa \\
       \addr University of Jeddah \\
       % % College of Computer Science and Engineering
       \AND
       \name Ateeq Ur Rehman Butt \email ateeqbutt13@live.com \\
        \addr National Textile University \\
       % Faisalabad, 37610, Pakistan
       % 3700 San Martin Drive,\\
       % Baltimore, MD 21218 USA
       % \AND
       % \name Philip Laird \email laird@ptolemy.arc.nasa.gov \\
       % \addr NASA Ames Research Centre,
       % AI Research Branch, Mail Stop: 269-2,
       % Moffett Field, CA  94035 USA
       }

% \author{\name Tariq Alsahfi \email tmalsahfi@uj.edu.sa\\
%        \addr Southwest Jiaotong University,\\
%        Chengdu Sichuan, 611756 China
%        % \AND
%        % \name Ateeq Ur Rehman Butt \email ateeqbutt13@live.com \\
%        % \addr National Textile University, \\
%        % Faisalabad, 37610, Pakistan
%        % 3700 San Martin Drive,\\
%        % Baltimore, MD 21218 USA
%        % \AND
%        % \name Philip Laird \email laird@ptolemy.arc.nasa.gov \\
%        % \addr NASA Ames Research Centre,
%        % AI Research Branch, Mail Stop: 269-2,
%        % Moffett Field, CA  94035 USA
%        }
% For research notes, remove the comment character in the line below.
% \researchnote

\maketitle

\begin{abstract}
Enhancing the reasoning capabilities of Large Language Models remains a critical challenge in artificial intelligence. We introduce RDoLT (Recursive Decomposition of Logical Thought) prompting, a novel framework that significantly boosts LLM reasoning performance. RDoLT is built on three key innovations: (1) recursively breaking down complex reasoning tasks into sub-tasks of progressive complexity; (2) employing an advanced selection and scoring mechanism to identify the most promising reasoning thoughts; and (3) integrating a knowledge propagation module that mimics human learning by keeping track of strong and weak thoughts for information propagation. Our approach was evaluated across multiple benchmarks, including GSM8K8, SVAMP, MultiArith, LastLetterConcatenation, and Gaokao2023 Math. The results demonstrate that RDoLT consistently outperforms existing state-of-the-art techniques, achieving a 90.98\% accuracy on GSM8K with ChatGPT-4 surpassing SOAT by 6.28\%. Similar improvements were observed on other benchmarks, with accuracy gains ranging from 5.5\% to 6.75\%. These findings highlight RDoLT’s potential to advance prompt engineering, offering a more effective and generalizable approach to complex reasoning tasks. 
% The code is publicly available on \href{https://github.com/Kaleemullahqasim/RDoLT_Prompting}{GitHub}.

\end{abstract}

\section{Introduction}
\label{Introduction}
Large Language Models (LLMs) have made significant strides in natural language understanding \cite{unlocking_the_potential_a_comprehensive_exploration_of_large_language_models_in_natural_language_processing,llm_fun_challenges_opp_and_social_impact} and text generation \cite{generative_ai_in_the_era_of_transformers_revolutionizing_natural_language_processing_with_llms,erdem2022neural}, enabling advancements in applications such as machine translation \cite{generative_ai_in_the_era_of_transformers_revolutionizing_natural_language_processing_with_llms,llm_based_machine_translation_for_ecommerce}, question-answering systems \cite{Q&A,Question-Asking_LLMs}, information retrieval \cite{information_retrieval_llms, inforRetmeetLLMs, van2024hybrid,erdem2022neural}, conversational agents \cite{agent_llms,xing2024designing, zhang2023exploring}, and text summarization \cite{text_Summary}. These models, trained on vast datasets, can generate human-like text \cite{LLMs_generate_human-like_opinions,llm_as_A_testing_expert-gui_testing} and produce sophisticated responses, positioning them as key players in industries such as healthcare, education, and legal services \cite{llm_healthcare,llm_and_Ai_reach_and_impact}. 
However, despite their achievement, LLMs still struggle with complex reasoning tasks \cite{large_language_models_are_not_strong_abstract_reasoners,large_language_models_cannot_self_correct_reasoning_yet,small_llms_are_weak_tool_learners_a_multi_llm_agent, lin-etal-2020-commongen}, such as mathematical problem-solving \cite{can_llm_master_math_no}, arithmetic, and common sense reasoning \cite{zhou2024misinformingllmsvulnerabilitieschallenges, lin-etal-2020-commongen}. These tasks require not only language comprehension but also the application of logical steps \cite{abstraction_of_thought_makes_language_models_better_reasoners}, deeper planning \cite{llm_task_planning}, and extensive thought exploration \cite{topologies_of_reasoning_demystifying_chains_trees_and_graphs_of_thoughts} to form consistent and accurate answers. LLMs often fall short in these areas, generating incorrect or hallucinated responses \cite{can_github_issues_be_solved_with_tree_of_thoughts_,chen2024inside}, which underscores the need for enhanced techniques to improve reasoning capabilities.

To address these challenges, prompting techniques \cite{vatsal2024surveypromptengineeringmethods} such as Chain-of-Thought (CoT) \cite{chain_of_thought_prompting_elicits_reasoning_in_large_language_models} have been developed to guide LLMs in generating intermediate reasoning steps before arriving at the final answer. CoT improves reasoning by generating intermediate solutions with “step-by-step” reasoning, thereby facilitating better logical progression. CoT has been further refined to Chain-of-Thought Self-consistency (CoT-SC) \cite{self_consistency_improves_chain_of_thought_reasoning_in_language_models}, which asks models to generate consistent thought chains and derive a final answer through majority voting of generated thoughts. Least-to-Most (L2M) \cite{least_to_most_prompting_enables_complex_reasoning_in_large_language_models} took a different approach to address complex problems by initially solving the simplest aspects, gradually progressing to the final solution of the task, where answers to simple aspects help the model understand the overall query. These methods have shown promise but still face significant limitations \cite{can_github_issues_be_solved_with_tree_of_thoughts_}, particularly in their thought selection and scoring for intermediate thoughts \cite{chain_of_thoughtlessness}.

\begin{figure}[ht]
    \centering
    \includegraphics[width=1\linewidth]{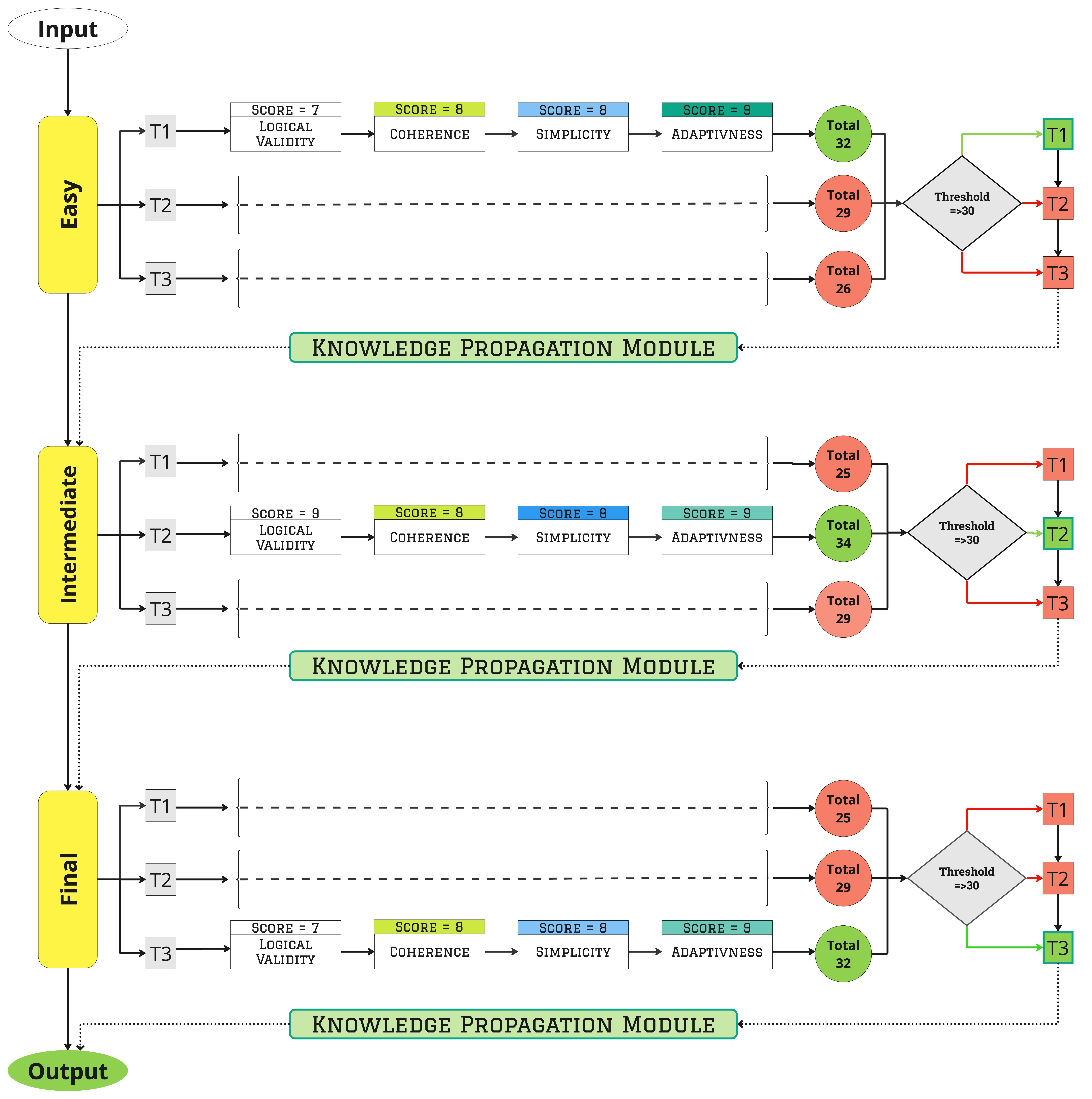}\caption{Illustration of Recursive Decomposition of Logical Thoughts (RDoLT) Framework. Each Yellow box represents decomposed task into Easy, Intermediate, and Final tiers. Each tier generates multiple thoughts (T1, T2, T3) evaluated on four features (logical validity, coherence, simplicity, and adaptiveness). Thoughts meeting the threshold criteria (green) are propagated to the next tier via the Knowledge Propagation Module (KPM), which tracks selected (strong) and rejected (weak) thoughts to inform future evaluations.}
    \label{fig:Framework-RDoLT}
\end{figure}

Existing techniques have significantly advanced the reasoning capabilities of LLMs, yet they continue to exhibit critical shortcomings when applied to complex reasoning tasks. CoT operates by generating a sequential reasoning process, breaking tasks into intermediate steps to arrive at a final answer. While this approach aids in structured reasoning, it suffers from a notable limitation: if an incorrect intermediate thought is generated, it propagates to subsequent steps, compounding errors and leading to inaccurate conclusions \cite{chain_of_thoughtlessness}. \cite{self_consistency_improves_chain_of_thought_reasoning_in_language_models} attempts to address this issue by generating multiple reasoning paths and using majority voting to determine the final answer. However, this method can be problematic when only two reasoning paths are generated or when the majority voting overlooks correct but rare outliers, leading to potentially sub-optimal decisions.

% \textcolor{red}{I think here you are start about the Least2Most method, just for clarity if you start with an statment like another method is Least2... method since the above about CoT} this is about Least2most as you can see in the cite
The design of \cite{least_to_most_prompting_enables_complex_reasoning_in_large_language_models}, which scaffolds learning by progressing from simple to more complex sub-tasks, faces the problem of error propagation. If incorrect conclusions are drawn in the initial simpler steps, these errors can cascade, impacting the overall solution. Moreover, these methods fail to fully integrate their respective strengths. For example, while CoT focuses on logical breakdown, it lacks the flexibility needed to adapt to real-time complexity, and Least2Most, despite its step-wise progression, struggles with scaling effectively to more nuanced problems. \cite{can_separators_improve_chain_of_thought_prompting_} introduces alternative approach that improves reasoning by adding separators between thoughts to encourage clearer delineation between reasoning stages, seeks to combine reasoning with action by having models alternate between reasoning steps and taking actions (such as querying external knowledge sources) during the thought process it mitigates some issues related to thought propagation, its reliance on external actions introduces new challenges, such as dependency on the availability of accurate external knowledge.

In response to these persistent limitations, we propose a novel prompting technique called Recursive Decomposition of Logical Thoughts (RDoLT) designed to address the key shortcomings of previous methods by integrating a more dynamic and recursive structure into the reasoning process. RDoLT enhances traditional CoT and CoT-SC methodologies by breaking down tasks into three distinct levels of complexity—easy, intermediate, and final while incorporating a robust thought evaluation and scoring system as shown in Fig\ref{fig:Framework-RDoLT}. Each stage generates multiple thoughts, assessed and scored based on four critical dimensional features: Logical Validity, Coherence, Simplicity, and Adaptiveness. These features allow RDoLT to evaluate thoughts not only based on their immediate correctness but also on their alignment with the overall task, their clarity, and their flexibility in different contexts.

Crucially, RDoLT introduces a Knowledge Propagation Module (KPM), a novel mechanism that tracks both selected and rejected thoughts throughout the reasoning process. By storing and propagating information about rejected thoughts (classified as “weak”), RDoLT ensures that potentially valuable ideas are not lost prematurely. This allows the system to revisit rejected thoughts when they become relevant in later stages of reasoning, minimizing the risk of missing out on correct but initially discarded solutions. Unlike previous approaches that discard non-majority reasoning paths, RDoLT’s KPM continuously refines its understanding by considering the full spectrum of thoughts generated.

Our contributions can be summarized through three primary innovations:
\circled{1} \textbf{Task Decomposition:} RDoLT decomposes reasoning tasks into three levels—easy, intermediate, and final—allowing for a more structured exploration of task complexity. This differs from the rigid structure of Least2Most prompting \cite{least_to_most_prompting_enables_complex_reasoning_in_large_language_models}, as RDoLT does not prioritize simpler tasks but instead decomposes tasks based on logical progressive complexity.
\circled{2} \textbf{Thought Scoring System:} RDoLT evaluates thoughts using a four-feature system—Logical Validity, Coherence, Simplicity, and Adaptiveness—ensuring that thoughts are selected not just for their correctness but also for their ability to support consistent and flexible reasoning across different contexts. \circled{3} \textbf{Knowledge Propagation Module (KPM):} The KPM tracks both successful (strong) and rejected (weak) thoughts, allowing for dynamic re-evaluation and adaptation throughout the reasoning process. This dual-path system mitigates the issues of premature rejection or over-reliance on majority voting seen in CoT-SC \cite{self_consistency_improves_chain_of_thought_reasoning_in_language_models}. 

Our research investigates the following key questions:
% \begin{enumerate}
%     \item \textbf{RQ1}: How does the RDoLT compare to existing prompting methods in terms of reasoning performance on complex tasks?
%     \item \textbf{RQ2}: What impact does the novel thought selection system have on the accuracy and efficiency of LLM reasoning?
%     \item \textbf{RQ3}: How effective is the Knowledge Propagation Module (KPM) in utilizing both selected and rejected thoughts to enhance reasoning outcomes?
%     \item \textbf{RQ4}: Can the decomposition strategy of RDoLT be generalized across different types of reasoning tasks and domains?
% \end{enumerate}
\begin{itemize}
    \item \textbf{RQ1}: How does the RDoLT compare to existing prompting methods in terms of reasoning performance on complex tasks?
    \item \textbf{RQ2}: What impact does the novel thought selection system have on the accuracy and efficiency of LLM reasoning?
    \item \textbf{RQ3}: How effective is the Knowledge Propagation Module (KPM) in utilizing both selected and rejected thoughts to enhance reasoning outcomes?
    % \item \textbf{RQ4}: Can the decomposition strategy of RDoLT be generalized across different types of reasoning tasks and domains?
\end{itemize}

Empirical evaluations demonstrate that RDoLT significantly improves LLM performance on a range of reasoning benchmarks, addressing key challenges identified in earlier prompting methods. By synthesizing insights from CoT, CoT-SC, Least2Most, and other techniques, RDoLT offers a comprehensive framework that enhances the logical reasoning capabilities of LLMs, providing a promising path forward in prompt engineering research. All codes and results will be open-sourced upon publication.

\section{Related Work}
\label{Relatedwork}
\begin{itemize}

\item \textbf{Feedback Guided Thought Generation:} Human feedback has been shown to enhance the performance of LLMs by providing an external evaluation that can refine model outputs \cite{learntorepair, NL-Edit, reinforcmentfromhumanfeedback}. However, this feedback is often expensive and problematic to incorporate into an automated process. Consequently, researchers have begun to replace human feedback with heuristic functions, which serve as a more scalable solution\cite{RainierReinforcedKnowledgeIntrospector, NEURIPS2022_b125999b, NEURIPS2022_8636419d, welleck2022generatingsequenceslearningselfcorrect}. 

Recent advancements have introduced self-reflective mechanisms where models generate their own feedback to assess and improve outputs \cite{madaan2023selfrefineiterativerefinementselffeedback, shinn2023reflexionlanguageagentsverbal, paul2024refinerreasoningfeedbackintermediate}. These techniques are especially beneficial for code generation and other multi-step tasks, as seen in \cite{chen2023teachinglargelanguagemodels}, which utilizes execution results for refinement. However, these approaches typically follow a linear left-to-right process, which limits exploration of alternative reasoning paths. In contrast, RDoLT introduces a broader, more flexible scoring-based feedback mechanism. Each thought node generates multiple child nodes, allowing exploration of alternative reasoning and improving decision-making comprehensiveness.

\item \textbf{Graph \& Tree-Based Reasoning:}  Tree-based approaches, such as the Tree of Thoughts (ToT) method, organize reasoning paths into a tree structure, allowing models to explore multiple decision branches \cite{tree_of_thoughts_deliberate_problem_solving_with_large_language_models, xie2023selfevaluationguidedbeamsearch}. These methods are particularly suited for multi-step problem-solving, where each node represents a partial solution. However, ToT’s rigid structure prohibits modification of intermediate nodes, which can result in a final solution that depends heavily on the initial steps. Once a branch is selected, there’s no opportunity for feedback or revision until the final answer is generated, these methods are also very cost-intensive if a long tree of thought is generated. 
Graph-based approaches, such as Graph of Thoughts (GoT), offer more flexibility by connecting reasoning steps as nodes within a graph, enabling multiple solution paths to be explored concurrently \cite{graph_of_thoughts_solving_elaborate_problems_with_large_language_models}. This flexibility is particularly beneficial for tasks with complex dependencies. However, the complexity of managing graph-based structures, especially for large tasks, presents significant computational challenges. 
% In comparison, RDoLT builds upon these approaches by allowing intermediate nodes to be revisited and modified. This flexibility introduces an essential feature in our framework: the Knowledge Propagation Module (KPM). The KPM ensures that selected and rejected thoughts from earlier stages are available for review in later stages. This dynamic feedback allows for iterative improvement, reducing the likelihood of error propagation across the reasoning chain and enabling a more nuanced exploration of the solution space.

\item \textbf{Thought Selection and Scoring Systems:}  Several methods have been developed to guide thought selection and scoring in LLM reasoning tasks, with each method offering a different approach to evaluating model outputs. CoT prompting has gained prominence by improving LLMs’ reasoning through the generation of intermediate reasoning steps \cite{chain-of-thought_prompting_elicits_reasoning_in_large_language_models}. While effective in breaking down multi-step problems, CoT treats the intermediate reasoning process as a “black box,” meaning that no evaluation occurs at each step before the final answer is generated. \cite{self_consistency_improves_chain_of_thought_reasoning_in_language_models} addresses this limitation by generating multiple reasoning paths and then marginalizing them to select the most consistent solution. This technique increases the likelihood of selecting the correct answer, however, CoT-SC improves upon CoT’s limitations, and it relies heavily on the assumption that majority voting across generated reasoning paths will produce an optimal solution. \cite{least_to_most_prompting_enables_complex_reasoning_in_large_language_models} on the other hand, adopts a hierarchical approach by first solving simpler problems before tackling more complex tasks. This incremental process builds model confidence; however, the decomposition is still treated as an automated “black box,” meaning that mistakes in early stages are not corrected before proceeding to more complex stages. 
% In contrast, RDoLT framework introduces a task complexity-based decomposition into three distinct levels: easy, intermediate, and final. Each level generates multiple candidate thoughts, which are evaluated based on specific features such as logical validity, coherence, simplicity, and adaptiveness. This feature-based scoring system ensures that only the most robust thoughts are propagated to the next level, minimizing error propagation and allowing for a more thorough review of reasoning at each stage. The KPM also enables dynamic feedback and revision of previously rejected thoughts, making the model more adaptable and ensuring higher accuracy in complex tasks.

\item \textbf{LM Planning and Structured Reasoning:}  Long-form content generation and complex problem-solving often require high-level planning and structured reasoning. Techniques like natural language outliners and schemas have proven effective for these tasks, guiding models to generate coherent, multistep outputs \cite{mirowski2022cowritingscreenplaystheatrescripts}. These methods have been successfully applied to diverse domains, including video games, fact-checking, housekeeping, and code optimization, where a clear plan or outline helps models structure their reasoning \cite{yao2023reactsynergizingreasoningacting, lin-etal-2020-commongen, wang2024describeexplainplanselect}. The challenge with these approaches is that while they provide structure, they do not necessarily allow for the revision of thoughts generated during the process. Once a plan is in place, the model tends to follow it rigidly, without reconsidering whether earlier decisions were correct or optimal. 
    
\end{itemize}

Our work addresses these limitations by allowing for both structured reasoning and flexible revisions. The decomposition of tasks based on task complexity into distinct levels—easy, intermediate, and final—ensures that the model progresses logically, but the feedback mechanism embedded within the KPM allows for real-time adjustments. If an intermediate thought proves incorrect or sub-optimal, the model can revise it without having to restart the entire reasoning process. This dynamic combination of planning and real-time evaluation offers a more robust approach to long-form and complex problem-solving.

\section{Methods}
\label{Methods}
The RDoLT employs a three-stage iterative reasoning process (\textit{Easy}, \textit{Intermediate}, \textit{Final}) to systematically refine outputs. At each stage, candidate thoughts (\(T_1, T_2, T_3,..T_n\)) are generated and evaluated using four scoring criteria: Logical Validity, Coherence, Simplicity, and Adaptiveness. Thoughts exceeding a predefined threshold are propagated to the next stage through the Knowledge Propagation Module (KPM), which integrates and refines selected outputs. This structured, score-driven approach ensures progressive enhancement of reasoning quality and convergence towards optimal solutions. The subsequent sections provide a detailed exposition of the framework’s stages, the scoring methodology, and the propagation mechanism.

\subsection{Task Decomposition}
The initial phase involve the decomposition of the reasoning task into three distinct levels based on gradual and progressive complexity: easy, intermediate, and final. This hierarchical decomposition is more sophisticated than the (\cite{least_to_most_prompting_enables_complex_reasoning_in_large_language_models}) by incorporating a more granular and human-level intelligent method of task segmentation. Given a complex reasoning task \( R \), we decompose it into three sub-tasks, \( P = \{R_{\text{easy}}, R_{\text{intermediate}}, R_{\text{final}}\} \). Each sub-task is designed to incrementally build upon the previous one, ensuring that the model tackles simpler components first and progressively moves to more complex reasoning.
The decomposition process can be represented as follows:
\begin{equation}
R_{\text{easy}} = f_{\text{decomp}}(R, \theta_{\text{easy}})
\end{equation}

\begin{equation}
R_{\text{intermediate}} = f_{\text{decomp}}(R, \theta_{\text{intermediate}})
\end{equation}

\begin{equation}
R_{\text{final}} = f_{\text{decomp}}(R, \theta_{\text{final}})
\end{equation}

% \[ t_{k,i} \sim p_{\theta}(t_{k,i} | I(t_{k,1}, t_{k,2}, \ldots, t_{k,i-1}, X, Q)) \]
The transition between these levels is not merely sequential but involves a feedback mechanism where the output of each level informs the subsequent level. This recursive feedback mechanism can be defined as:

\begin{equation}
t_{k+1,i} = f_{\text{feedback}}(t_{k,i}, \theta_{k+1})
\end{equation}

where \( k \) represents the current level. This feedback mechanism integrates the output of the \( k \)-th level to refine the input for the \( (k+1) \)-th level, ensuring that knowledge and errors identified in earlier stages are propagated and corrected in later stages.

This approach significantly reduces cognitive overload on the model and mirrors the step-by-step approach humans naturally employ when solving complex problems. By systematically refining thoughts and leveraging a robust scoring system, the RDoLT framework enhances the reasoning performance of LLMs. This method is not only more nuanced but also more aligned with human cognitive processes\cite{decomosition_of_task_with_llms}, thereby improving the model’s accuracy and consistency in solving complex reasoning tasks.

\subsection{Thought Generation}
Thought generation is a critical component that occurs within each of the decomposition levels: easy, intermediate, and final. The process involves generating multiple candidate thoughts for each task segment to ensure a diverse set of potential solutions is explored. For our framework, we set \( n \) (the number of thoughts generated per level) to three. 
Given a decomposed task \( R_k \) at level \( k \), the thought generation process aims to produce a set of candidate thoughts \( T_k = \{t_{k1}, t_{k2}, t_{k3}\} \). Each thought is generated by the LLM based on the input \( X \), the question \( Q \), and any previously generated thoughts at that level.
Formally, the thought generation process at level \( k \) is represented as:
\begin{equation}
t_{ki} \sim p_{\theta}(t_{ki} \mid I(t_{k1}, t_{k2}, \ldots, t_{k(i-1)}, X, Q)), \quad \text{for } i = 1, 2, 3
\end{equation}

where \( p_{\theta} \) denotes the probability distribution parameterized by \(\theta\), and \( I(\cdot) \) indicates that the prompt includes all previous thoughts, task instructions \( X \), and the corresponding question \( Q \). 
The thoughts generated at each level \( T_E = \{t_{E1}, t_{E2}, t_{E3}\} \), \( T_I = \{t_{I1}, t_{I2}, t_{I3}\} \), and \( T_F = \{t_{F1}, t_{F2}, t_{F3}\} \) undergo evaluation based on predefined criteria in the subsequent scoring system step.

\subsection{Scoring and Evaluation}
The scoring system in our framework evaluates each generated thought at each decomposition level using four core features: \textbf{Logical Validity}, \textbf{Coherence}, \textbf{Simplicity}, and \textbf{Adaptiveness}. These features ensure that the selected thoughts are effective for reasoning and reflect human-like intelligence. 

\begin{itemize}
    \item \textbf{Logical Validity} (\(S_{\text{valid}}\)) ensures the thought is logically sound. This can be represented as the negative penalty based on logical contradictions:
    \begin{equation}
    S_{\text{valid}}(t_{ki}) = -\sum_{r \in \text{rules}} \mathbb{I}\{\text{violates}(r, t_{ki})\}
    \end{equation}
    where \(\mathbb{I}\) is an indicator function that is 1 if the thought \(t_{ki}\) violates a known logical rule \(r\), and 0 otherwise. A lower score penalizes thoughts that violate known rules or facts. Human or model can both be used as a scorer of all these following steps.
    
    \item \textbf{Coherence} (\(S_{\text{cohere}}\)) measures the degree to which the thought follows from previous thoughts, and can be defined as a similarity measure between the current thought and previous thoughts:
    \begin{equation}
    S_{\text{cohere}}(t_{ki} \mid \{t_{k1}, t_{k2}, \ldots, t_{k(i-1)}\}) = \frac{1}{i-1} \sum_{j=1}^{i-1} \text{sim}(t_{ki}, t_{kj})
    \end{equation}
    
    where \(\text{sim}(\cdot)\) represents a similarity function (cosine similarity) between the current thought \(t_{ki}\) and process thoughts \(t_{kj}\). Sentences similarity can be calculated using any embedding or sentence transformer model; however to reduce the complexity of the whole proces we let LLM score the thought \cite{large_language_models_are_contrastive_reasoners}.
    
    \item \textbf{Simplicity} (\(S_{\text{simple}}\)) evaluates the clarity and conciseness of a thought, inversely related to its complexity. This can be modeled by penalizing the length or complexity of the thought:
    \begin{equation}
    S_{\text{simple}}(t_{ki}) = -\text{complexity}(t_{ki})
    \end{equation}
    where \(\text{complexity}(t_{ki})\) could represent the length of the thought or the number of steps in reasoning, with lower complexity resulting in a higher score.
    
    \item \textbf{Adaptiveness} (\(S_{\text{adapt}}\)) assesses how well the thought aligns with the external context, such as the task instructions \(X\) and the question \(Q\). It can be defined as:
    \begin{equation}
    S_{\text{adapt}}(t_{ki} \mid \{X, Q\}) = \text{sim}(t_{ki}, \{X, Q\})
    \end{equation}
    where \(\text{sim}(\cdot)\) measures the similarity between the thought and the context provided by \(X\) and \(Q\).
\end{itemize} 

The overall score for each thought \( t_{ki} \) is the sum of its individual feature scores:
\begin{equation}
S(t_{ki}) = S_{\text{valid}}(t_{ki}) + S_{\text{cohere}}(t_{ki}) + S_{\text{simple}}(t_{ki}) + S_{\text{adapt}}(t_{ki})
\end{equation} 
Thought is selected if its total score exceeds a predefined threshold \(\tau\) , or if it maximizes the score among all thoughts in the current decomposition level: 
\begin{equation}
t_{k}^{*} = \arg\max_{t_{ki}} S(t_{ki})
\end{equation} 

\begin{figure}[htbp]
    \centering
    \includegraphics[width=1\linewidth]{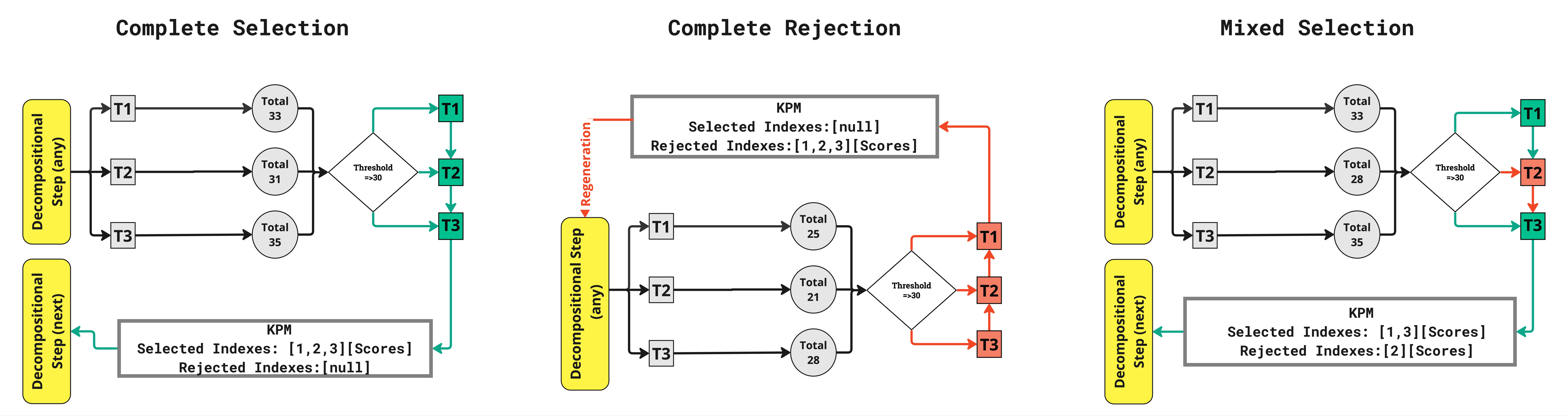}
    \caption{Edge cases handled by the Knowledge Propagation Module (KPM) during thought selection in decomposition steps. The complete Selection scenario (left) shows all thoughts (T1, T2, T3) selected. The Complete Rejection scenario (center) depicts all thoughts rejected, leading to regeneration. Mixed Selection scenario (right) highlights partial selection, with some thoughts accepted and others rejected. These scenarios demonstrate how the KPM ensures optimal thought progression}
    \label{fig:edge-case}
\end{figure}

In contrast to \cite{chain_of_thought_prompting_elicits_reasoning_in_large_language_models, least_to_most_prompting_enables_complex_reasoning_in_large_language_models, tree_of_thoughts_deliberate_problem_solving_with_large_language_models, graph_of_thoughts_solving_elaborate_problems_with_large_language_models, algorithm_of_thoughts_enhancing_exploration_of_ideas_in_large_language_models}, our framework employs a systematic, feature-based evaluation and scoring system of thoughts at each step, ensuring not only the highest-quality thoughts progress but also maintaining knowledge of weak thoughts to feed the model in subsequent steps of decomposition. This recursive, detailed scoring process—based on logical validity, coherence, simplicity, and adaptiveness—enables a more nuanced and reliable reasoning mechanism. 

\subsection{Knowledge Propagation Module and Edge Case Management}

The Knowledge Propagation Module (KPM) plays a crucial role in the RDoLT framework’s reasoning process. It is responsible for managing knowledge and propagating it through the subsequent steps of reasoning. This module ensures that the flow of information remains coherent and consistent across all levels of decomposition, significantly enhancing the model's reasoning capabilities. Furthermore, the KPM manages the execution of the system, handles edge cases, and oversees the selection and rejection of thoughts, which is essential for maintaining the framework's overall accuracy.

\begin{figure}[ht]
    \centering
    \includegraphics[width=1\linewidth]{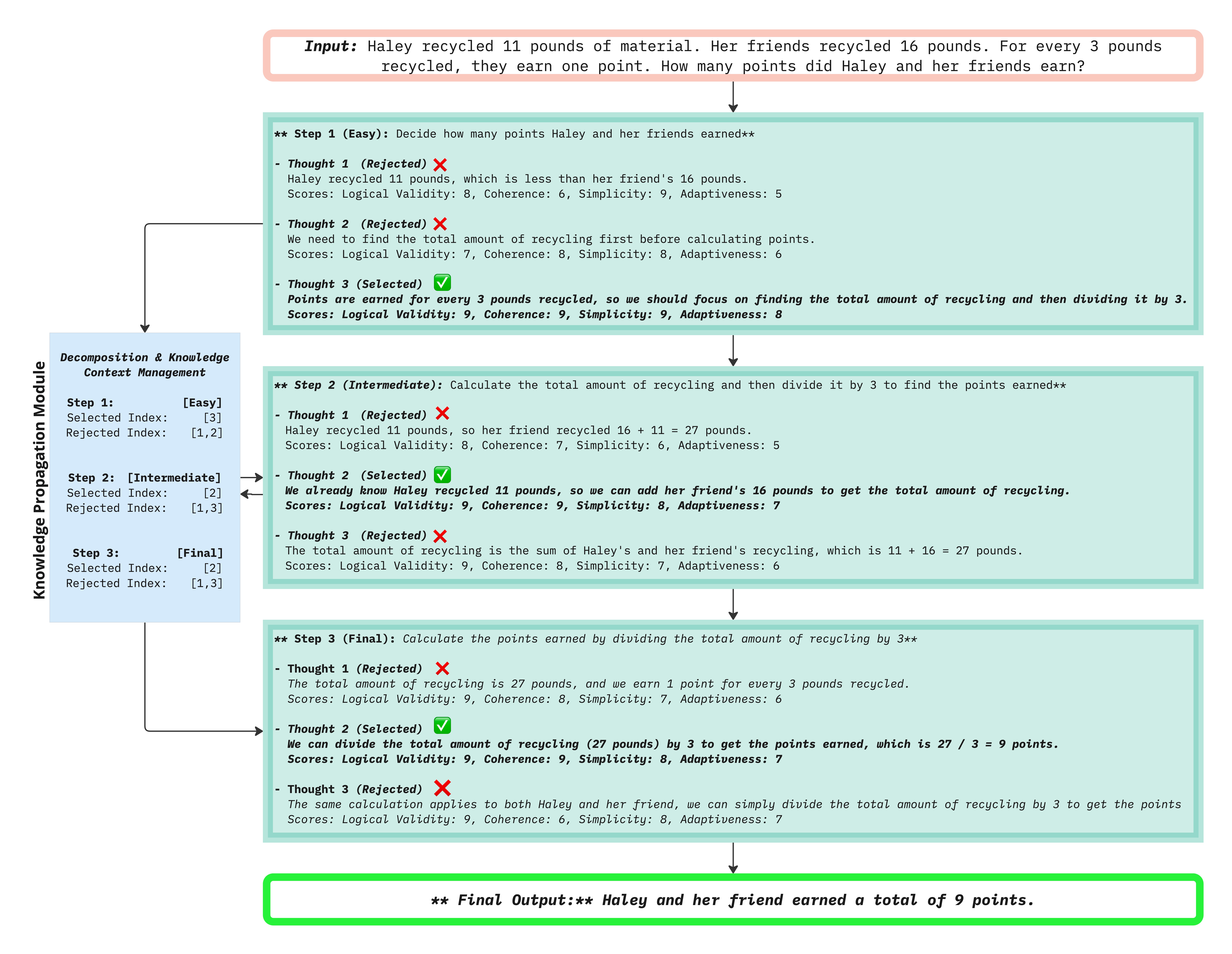}
    \caption{Detailed example illustrating how RDoLT addresses problems by \protect\circled{1} decomposing them into three reasoning steps, \protect\circled{2} generating thoughts for each step, and \protect\circled{3} scoring and propagating knowledge through a Knowledge Propagation Module (KPM) across subsequent steps. The system performs scoring and selection at each step, ensuring that both accepted and rejected thoughts are transmitted to enhance learning and understand why certain thoughts were discarded}
    \label{fig:example}
\end{figure}

The KPM tracks both selected and non-selected thoughts at each step of the reasoning process. Selected thoughts are those that have met the threshold criteria based on the scoring system, while non-selected thoughts (weak thoughts) are those that did not meet the required threshold. Unlike traditional methods, which focus primarily on the immediate next step, KPM makes this information available to all subsequent steps. For instance, thoughts selected or rejected during the easy step are accessible in both the intermediate and final steps. This comprehensive tracking ensures that the system retains a full understanding of the reasoning progression from start to finish.

Mathematically, let \( S_{\text{selected}}^k \) and \( S_{\text{non-selected}}^k \) represent the sets of selected and non-selected thoughts at level \( k \), respectively. The KPM propagates these sets to all subsequent levels \( k+1, k+2, \ldots \), as follows:
\begin{equation}
\{S_{\text{selected}}^k, S_{\text{non-selected}}^k\} \rightarrow \{S_{\text{selected}}^{k+1}, S_{\text{non-selected}}^{k+1}, \ldots, S_{\text{selected}}^{n}, S_{\text{non-selected}}^{n}\}
\end{equation}

This propagation includes maintaining a history of all thoughts and providing this history to the reasoning framework to ensure well-informed decision-making at each step. Additionally, the KPM includes a robust feedback mechanism. If no thought passes the threshold at any step, the module informs the main framework to regenerate thoughts, ensuring the reasoning process does not stall. This feedback mechanism is critical for maintaining the flow of reasoning and preventing bottlenecks:
\begin{equation}
\text{If } S_{\text{selected}}^k = \emptyset \Rightarrow \text{Regenerate thoughts at level } k
\end{equation}

Moreover, the KPM handles various edge cases, such as those illustrated in Figure 3. It tracks thoughts that receive the same score and ensures appropriate handling. For instance, if multiple thoughts pass the threshold, the module informs the model of the scores for all passing thoughts, enabling it to prioritize them effectively. If two thoughts have identical scores, they are given equal priority and propagated to the next step:
\begin{equation}
\text{If } S(t_{k1}) = S(t_{k2}) \text{ and both } t_{k1}, t_{k2} \in S_{\text{selected}}^k \Rightarrow t_{k1}, t_{k2} 
\end{equation}

In cases where all thoughts pass the threshold, the module provides detailed scores to help the model utilize the thought information more effectively in subsequent steps. This systematic approach ensures that the reasoning process remains flexible and robust, capable of handling various scenarios without compromising the integrity of the reasoning flow. Compared to CoT, CoT-SC, and Least2Most prompting, our KPM offers a more advanced and comprehensive approach to managing and propagating knowledge. Traditional methods primarily focus on sequential thought generation and consensus mechanisms without maintaining a detailed history of thoughts or providing robust feedback. Our KPM addresses these gaps by ensuring that all relevant information is available at every step, thereby enhancing the overall accuracy and reliability of the reasoning process.

\section{Experiments}
To evaluate the effectiveness of the RDoLT, we conducted a comprehensive series of experiments. These experiments aimed to assess RDoLT's performance across various reasoning benchmarks, comparing it with state-of-the-art prompting methods. We meticulously selected benchmarks and models, implemented comparative methodologies, and analyzed the main results to determine the robustness of different RDoLT variants. Additionally, we examined the impact of the quantity of thoughts generated on success rates to understand how thought granularity influences overall performance. The following sections provide a detailed account of our experimental setup and findings.

\subsection{Benchmarks \& Models Selection }
Given the versatility and ease of use offered by the RDoLT framework, as well as the accessibility of tools like LM-Studio\cite{lm_studio_-_discover_download_and_run_local_llms} and Ollama\cite{ollama}, we conducted experiments using four distinct open-source and open-weight LLMs: Llama-3 (8B)\cite{llama3modelcard}, QWEN-2 (7B)\cite{qwen2}, Gemma-2(9B)\cite{gemma_2024}, and Gemma-2(27B)\cite{gemma_2024}. Although we also utilized the OpenAI API to access ChatGPT-4o\cite{openai2024gpt4technicalreport}, our primary focus was on evaluating the performance of open-source LLMs with various quantization levels. This focus allowed us to conduct extensive experiments and explore different variations in prompt design.

In our experiments, we maintained the temperature parameter at 0.4, striking a balance between encouraging model creativity and ensuring consistent, reliable responses for complex reasoning tasks. Additionally, the context length was set at 8192 tokens to maximize the model’s ability to handle extensive input sequences. We also explored context lengths of 4096 and 2056 tokens to evaluate the impact on model performance and accuracy.

The RDoLT framework is specifically designed to address reasoning tasks, particularly those requiring sequential and multi-step reasoning. To thoroughly evaluate the effectiveness of our system, we tested it against well-known benchmarks that push the limits of prompt engineering. For mathematical reasoning, we deployed the GSMK8 \cite{gmsk8}, Multi-Arithmetic \cite{MultiArith}, SVAMP \cite{patel-etal-2021-nlp}  and Gaokao 2023 Math \cite{gaokao-math} benchmarks. To assess the system’s ability to handle common-sense reasoning, we included the LastLetterConcatenation \cite{LastLetterConcat} benchmark in our evaluation. In total, we tested our prompt design across five different benchmarks to assess its generalizability and compare its performance with state-of-the-art techniques.

\subsection{Method Selected for Comparison}
We compare our system with Standard I/O, (CoT), (CoT-SC). (Auto CoT) and (L2M)  prompting. These methods were chosen for their prominence and varied approaches to improving prompt accuracy. Standard I/O prompting serves as a baseline to highlight the improvements of more advanced techniques. CoT and CoT-SC and its other variants, which follow a step-by-step reasoning structure, align with our decomposition-based approach, while L2M offers a contrasting progressive complexity strategy. Unlike fine-tuning methods, which often excel in domain-specific tasks but lack flexibility and generalization across diverse datasets, RDoLT maintains adaptability. Evaluating RDoLT against these methods allows for a comprehensive assessment of its accuracy and generalizability.

\section{Main Results}
\begin{table*}[ht]
\centering
\caption{ Comprehensive evaluation of the RDoLT framework across various LLMs. The performance of RDoLT is compared with several established methodologies, including Vanilla prompting, CoT, CoT-SC, Least-to-Most prompting, and Auto-CoT
(A-CoT). The assessment encompasses multiple benchmarks: GSM8K, SVAMP, MultiArith, Last Letter Concatenation, and Gaokao 2023 Math.}
\label{tab:performance}
\begin{adjustbox}{max width=\textwidth}
\begin{tabular}{llcccccc}
\toprule
\textbf{Benchmark} & \textbf{Model} & \textbf{Vanilla} & \textbf{CoT} & \textbf{CoT-SC} & \textbf{Least2Most} & \textbf{Auto-CoT} & \textbf{RDoLT} \\ 
\midrule
\multirow{6}{*}{\begin{tabular}{@{}l@{}}GSM8K \\ \cite{gmsk8}\end{tabular}} 
 & ChatGPT 4o & 84.7 & 88.9 & 89.4 & - & 85.8 & \textbf{90.98}$\uparrow$ \\ 
 & LLama 3 (7B) & 67.42 & 71.29 & \textbf{72.86}$\uparrow$ & 69.91 & 68.53 & 72.63 \\
 & Qwen 2 (7B) & 62.2 & 65.3 & 64.8 & 63.6 & 61.8 & \textbf{67.4}$\uparrow$ \\
 & Gemma 2 (9B) & 59.36 & 64.28 & 62.53 & 61.89 & 60.18 & \textbf{65.79}$\uparrow$ \\
 & Gemma 2 (27B) & 60.65 & 75.46 & \textbf{76.72}$\uparrow$ & 74.94 & 71.83 & 76.58 \\ 
\midrule
\multirow{6}{*}{\begin{tabular}{@{}l@{}}SVAMP \\ \cite{patel-etal-2021-nlp}\end{tabular}} 
 & ChatGPT 4o & 83.5 & 87.3 & 86.7 & 85.9 & - & \textbf{89.35}$\uparrow$ \\
 & LLama 3 (7B) & 65.79 & \textbf{69.54}$\uparrow$ & 68.86 & 67.93 & 66.67 & 69.23 \\
 & Qwen 2 (7B) & 60.3 & 64.2 & 63.5 & 62.6 & 61.3 & \textbf{66.37}$\uparrow$ \\
 & Gemma 2 (9B) & 58.6 & 63.4 & 61.8 & 60.9 & \textbf{64.52}$\uparrow$ & 64.19 \\
 & Gemma 2 (27B) & 69.86 & 73.75 & 72.94 & 71.98 & 70.69 & \textbf{75.27}$\uparrow$ \\ 
\midrule
\multirow{6}{*}{\begin{tabular}{@{}l@{}}MultiArith \\ \cite{MultiArith}\end{tabular}} 
 & ChatGPT 4o & 82.7 & - & 85.7 & 84.9 & 83.6 & \textbf{88.2}$\uparrow$ \\
 & LLama 3 (7B) & 77.93 & 81.62 & 80.97 & 79.84 & 78.53 & \textbf{83.31}$\uparrow$ \\
 & Qwen 2 (7B) & 58.6 & 62.3 & 61.5 & 60.4 & 59.2 & \textbf{64.5}$\uparrow$ \\
 & Gemma 2 (9B) & 56.87 & 61.38 & 59.76 & 58.69 & 57.48 & \textbf{63.83}$\uparrow$ \\
 & Gemma 2 (27B) & 67.96 & \textbf{72.73}$\uparrow$ & 71.86 & 70.75 & 68.68 & 72.49 \\ 
\midrule
\multirow{6}{*}{\begin{tabular}{@{}l@{}}LastLetterConcatenation \\ \cite{LastLetterConcat}\end{tabular}} 
 & ChatGPT 4o & 80.4 & 84.4 & 83.7 & - & 81.3 & \textbf{87.15}$\uparrow$ \\
 & LLama 3 (7B) & 62.58 & 66.36 & 65.67 & 64.48 & 63.29 & \textbf{68.24}$\uparrow$ \\
 & Qwen 2 (7B) & 57.4 & 61.2 & 60.5 & 59.3 & 58.2 & \textbf{63.35}$\uparrow$ \\
 & Gemma 2 (9B) & 56.46 & 60.28 & 58.69 & 57.59 & 55.87 & \textbf{62.76}$\uparrow$ \\
 & Gemma 2 (27B) & 66.67 & 70.48 & 69.58 & 68.39 & 67.29 & \textbf{72.93}$\uparrow$ \\ 
\midrule
\multirow{6}{*}{\begin{tabular}{@{}l@{}}Gaokao 2023 Math \\ \cite{gaokao-math}\end{tabular}} 
 & ChatGPT 4o & - & 83.2 & 82.5 & 81.6 & 80.3 & \textbf{85.64}$\uparrow$ \\
 & LLama 3 (7B) & 60.86 & 64.49 & 63.78 & 62.68 & 61.39 & \textbf{66.57}$\uparrow$ \\
 & Qwen 2 (7B) & 55.7 & 59.4 & 58.6 & 57.5 & 56.3 & \textbf{61.75}$\uparrow$ \\
 & Gemma 2 (9B) & 55.27 & 59.19 & 57.58 & 56.47 & 54.79 & \textbf{61.68}$\uparrow$ \\
 & Gemma 2 (27B) & 65.47 & \textbf{70.28}$\uparrow$ & 69.39 & 68.26 & 66.18 & 70.05 \\ 
\bottomrule
\end{tabular}
\end{adjustbox}
\end{table*}

We evaluated the RDoLT framework on five benchmarks—\cite{gmsk8, LastLetterConcat,gaokao-math, MultiArith, patel-etal-2021-nlp} comparing it with established prompting techniques, including CoT, CoT-SC, Least2Most, and Auto-CoT (A-CoT). The results, shown in (Table \ref{tab:performance}), demonstrate that RDoLT consistently outperforms these methods across multiple tasks and models.

On the GSM8K benchmark, RDoLT achieves an accuracy of 90.98\% with ChatGPT-4o, surpassing CoT (89.4\%) and A-CoT (85.8\%). RDoLT also shows strong performance with LLama3 (72.63\%) and Qwen2 (67.4\%), indicating its adaptability across different models and architectures. For the SVAMP benchmark, RDoLT leads with 89.35\% using ChatGPT-4o, outperforming CoT-SC and CoT. CoT performs relatively well with LLama3 (69.54\%), indicating that certain models can benefit from specific prompting strategies, although RDoLT remains superior overall across various architectures. In the MultiArith benchmark, RDoLT achieves 88.2\% with ChatGPT-4o, showing strong results in arithmetic reasoning tasks. LLama3 (83.31\%) and Qwen2 (64.5\%) also perform well with RDoLT. However, in the case of the Gemma2 (27B) model, CoT produces better results, suggesting that the optimal prompting technique can vary depending on the model and task complexity. The Last Letter Concatenation benchmark further confirms RDoLT’s effectiveness, with ChatGPT-4o reaching 87.15\%. LLama3 (68.24\%) and Qwen2 (63.35\%) also perform well with RDoLT, reinforcing its generalization capability across different models and tasks. Finally, in the Gaokao 2023 Math benchmark, RDoLT achieves 85.64\% with ChatGPT-4o. LLama3 (66.57\%) and Qwen2 (61.75\%) also show strong results. While CoT slightly outperforms RDoLT with Gemma2 (70.28\%), RDoLT remains the most consistent top performer overall.

In summary, RDoLT outperforms other prompting methods in 65\% of the evaluated benchmarks. While CoT and CoT-SC demonstrate competitive results in specific cases, RDoLT offers the most consistent and robust performance across a diverse range of reasoning tasks and models.

\label{main-results}

\subsection{Robustness of RDoLT Variants}
\begin{table}[htbp]
\centering
\caption{Performance Analysis of RDoLT Variants Across Different Threshold Score Levels: the performance of six variants across four Threshold levels. Bold figures with upward arrows (↑) indicate peak performance for each variant. Results show that optimal thresholds vary among variants, with most peaking at 30 or 35.}
\label{tab:performance_variants}
\begin{adjustbox}{max width=\textwidth}
\begin{tabular}{@{}lcccc@{}}
\toprule
\textbf{Variants} & \textbf{Threshold $\geq$ 25} & \textbf{Threshold $\geq$ 30} & \textbf{Threshold $\geq$ 35} & \textbf{Threshold $\geq$ 40} \\ \midrule
Single-Step(Sequential)           & 53.12                 & 73.47                & \textbf{80.78}$\uparrow$        & 65.30     \\
% Multi-Step           & 60.24                 & \textbf{77.51}$\uparrow$       & 75.12                 & 73.54                \\
Single-Step(One-Shot)             & 68.67                 & \textbf{71.73}$\uparrow$       & 68.89                 & 69.45                \\
Few-Shots(2)            & 65.89                 & 76.47                & \textbf{77.15}$\uparrow$        & 70.60                \\
Multi-Requests(3)      & 61.12                 & \textbf{73.24}$\uparrow$       & 72.34                 & 74.15                \\
Multi-Requests Unlimited* & 60.24              & \textbf{74.51}$\uparrow$       & 73.13                 & 72.30                \\ \bottomrule
\end{tabular}
\end{adjustbox}
\end{table}

RDoLT variants across different threshold score levels revealed intriguing patterns, each with potential implications for practical applications. The single-step (sequential) variant demonstrated the highest overall performance, peaking at 80.78\% with a threshold of $\geq$ 35. This superior performance suggests that for complex tasks, allowing more extensive processing before making decisions yields better results. The sequential nature of this variant likely enables a more thorough exploration of the problem space, leading to more accurate solutions. Multi-Step (1-Shot) and Few-Shots (2) variants showed optimal performance at thresholds of $\geq$ 35 (77.51\%) and $\geq$ 35 (77.15\%), respectively. These results indicate that these approaches benefit from moderate thresholds, striking a balance between depth of processing and efficiency. The slightly lower performance compared to the Single-Step variant might be attributed to the trade-off between speed and accuracy, where these methods attempt to reach solutions more quickly but potentially at the cost of some precision. 

Interestingly, One-Shot variant achieved its best performance (71.73\%) at a lower threshold of $\geq$30. This suggests that this approach is more suitable for tasks requiring quicker decision-making or where rapid responses are valued over absolute accuracy. The lower overall performance compared to other variants implies while efficient, this method may sacrifice some problem-solving depth. Multi-Request variants, both limited (3) and unlimited, performed optimally at the $\geq$ 30 threshold, with scores of 73.24\% and 74.51\% respectively. This pattern indicates that allowing multiple attempts at problem-solving can be effective, but excessive iterations may not yield significant improvements. The similarity in performance between the limited and unlimited variants suggests that there might be a natural ceiling to the benefits of multiple attempts, beyond which additional requests do not substantially enhance outcomes.

A noteworthy trend across all variants is the reduced performance at the highest threshold ($\geq$ 40). This consistent drop-off points to a potential over-processing effect, where excessively high thresholds may introduce unnecessary complexity or lead to over-fitting in the decision-making process. This observation underscores the importance of finding the right balance in threshold setting to optimize performance without incurring diminishing returns.

\begin{table}[htbp]
\centering
\caption{Step-wise problem-solving performance across different step levels (Easy, Intermediate, Final) using varying numbers of thoughts per step (3, 5, and 7). Results show the progression of solved problems at each thought step, the total problems solved, and the corresponding success rates for each configuration. The 3 Thoughts Count/Step} configuration is identified as the best-performing system.
\label{tab:stepwise_analysis}
\begin{adjustbox}{max width=\textwidth}
\begin{tabular}{@{}ccccccccccc@{}}
\toprule
\textbf{Thoughts Count/Step} & \textbf{Steps} & \textbf{T1} & \textbf{T2} & \textbf{T3} & \textbf{T4} & \textbf{T5} & \textbf{T6} & \textbf{T7} & \textbf{Total Solved} & \textbf{Success Rate\%} \\ \midrule
\multirow{4}{*}{\textbf{3}} & Easy          & 2           & 3           & 4           & -           & -           & -           & -           & 9                      & 30                  \\
                            & Intermediate  & 4           & 5           & 6           & -           & -           & -           & -           & 15                     & 50                  \\
                            & Final         & 7           & 8           & 9           & -           & -           & -           & -           & 24                     & 80                  \\
                            & Total         & 13          & 16          & 19          & -           & -           & -           & -           & 48                     &\textbf{53.33}$\uparrow$         \\ \midrule
\multirow{4}{*}{5}          & Easy          & 2           & 3           & 4           & 5           & 6           & -           & -           & 20                     & 33.33                  \\
                            & Intermediate  & 4           & 5           & 6           & 7           & 8           & -           & -           & 30                     & 50                 \\
                            & Final         & 6           & 7           & 8           & 9           & 10          & -           & -           & 40                     & 66.67                  \\
                            & Total         & 12          & 15          & 18          & 21          & 24          & -           & -           & 90                     & \textbf{49.75}$\downarrow$                \\ \midrule
\multirow{4}{*}{7}          & Easy          & 1           & 2           & 3           & 2           & 4           & 4           & -           & 16                     & 31.11                  \\
                            & Intermediate  & 2           & 3           & 3           & 4           & 5           & 5           & 5           & 27                     & 38.89                  \\
                            & Final         & 1           & 4           & 6           & 5           & 6           & 6           & 6           & 34                     & 46.67                  \\
                            & Total         & 6           & 9           & 10          & 12          & 14          & 15          & 11          & 77                     & \textbf{38.89}$\downarrow$                \\ \bottomrule
\end{tabular}
\end{adjustbox}
\end{table}

\subsection{Impact of Thought Quantity on Success Rates}

\begin{figure}
    \centering
    \includegraphics[width=0.75\linewidth]{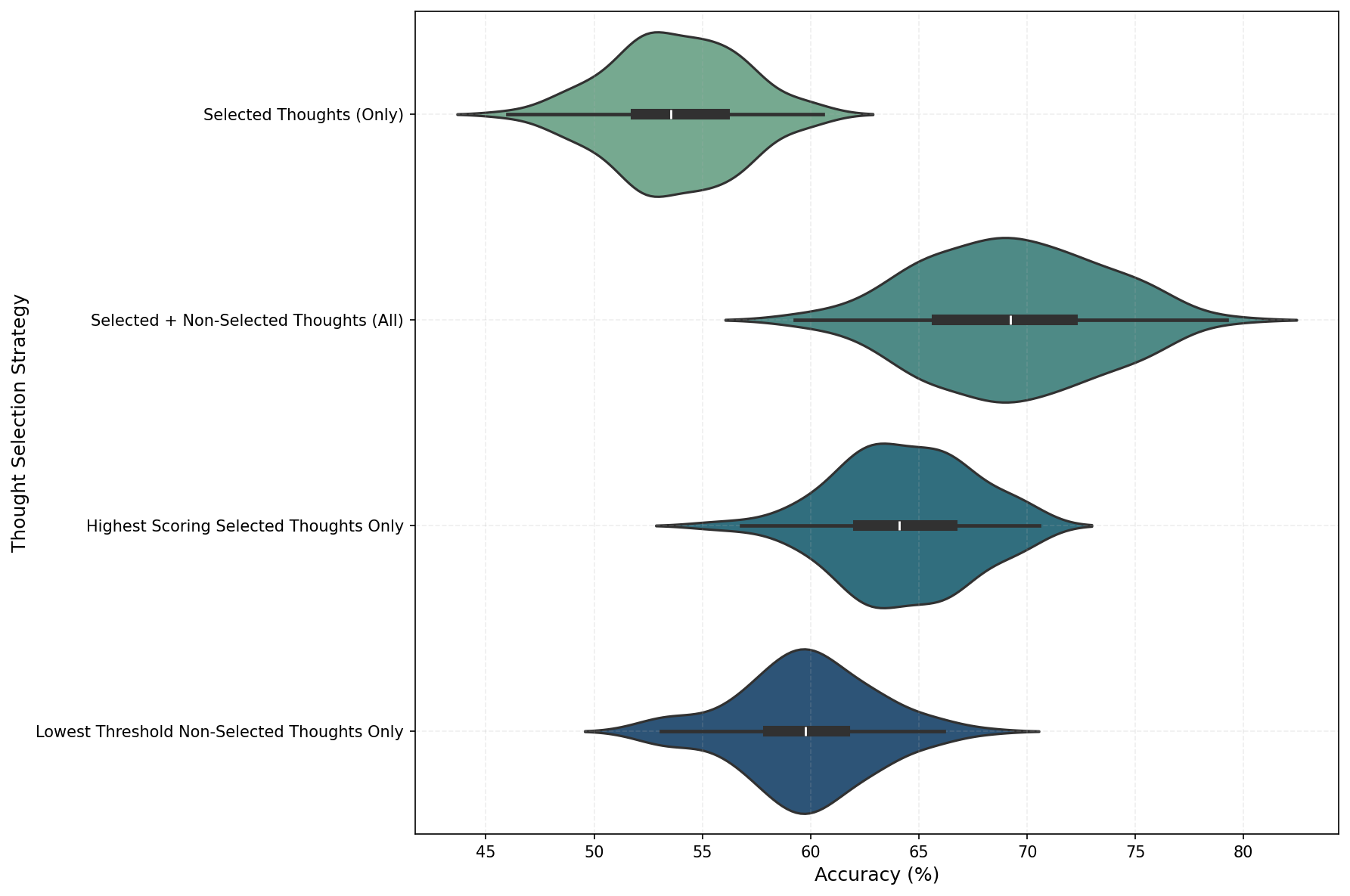}
    \caption{Impact of different thought selection strategies on accuracy using KPM. Strategies include using only selected thoughts, both selected and non-selected thoughts, highest-scoring selected thoughts, and lowest-threshold non-selected thoughts. Incorporating both selected and non-selected thoughts yields a wider range of accuracy, while selecting only final thoughts ensures more consistent accuracy results}
    \label{fig:ThougtQuality}
\end{figure}

The evaluation of the RDoLT prompt system reveals insightful findings about its performance across different stages of problem-solving, emphasizing the number of thoughts generated per step. Table 4 encapsulates the results of this analysis, illustrating the effectiveness of generating 3, 5, and 7 thoughts per step.

In scenarios where three thoughts were generated per step, the final step proved to be the most effective, solving 24 out of 30 problems, resulting in an 80\% success rate. This demonstrates the crucial role of the final step, benefiting from the refined thoughts developed in earlier stages. The intermediate step followed, solving 15 problems with a 50\% success rate, while the easy step had the lowest contribution, solving 9 problems with a 30\% success rate. Overall, this variant solved 48 problems with a success rate of 53.33\%, showcasing its robustness in a structured problem-solving environment.

When the number of thoughts per step increased to five, the final step maintained its significance, solving 40 problems with a 66.67\% success rate. The intermediate step solved 30 problems, achieving a 50\% success rate, while the easy step solved 20 problems, with a 33.33\% success rate. This variant solved a total of 90 problems, yielding a 60.00\% success rate. These results suggest that increasing the number of thoughts per step enhances problem-solving capabilities, though the final step remains critical for achieving high accuracy.

Conversely, generating seven thoughts per step showed a decrease in overall performance. The final step solved 34 problems, achieving a 46.67\% success rate, the intermediate step solved 27 problems with a 38.89\% success rate, and the easy step solved 16 problems with a 31.11\% success rate. The total number of problems solved in this variant was 77, with a 36.67\% success rate. This indicates that while more thoughts per step provide more options, it may also introduce complexity that can hinder overall effectiveness.

% \subsubsection{Selective Knowledge Propagation}

\subsection{Limitations and Future Directions}

While the RDoLT framework demonstrates superior performance across multiple benchmarks, several limitations remain. First, the generalizability of RDoLT to domain-specific tasks has not been fully explored. The framework shows promising results in standard reasoning tasks, but its adaptability to highly specialized domains such as legal reasoning or medical diagnostics may require further fine-tuning and optimization. Additionally, the computational overhead of maintaining the Knowledge Propagation Module (KPM) may limit scalability, particularly in resource-constrained environments.

Another potential threat to validity is the reliance on benchmark datasets that may not fully capture the complexity of real-world reasoning scenarios. Although benchmarks like GSM8K and SVAMP are widely used, they represent structured tasks that may not account for the diverse and dynamic nature of human reasoning in more unstructured settings.

Future research should focus on addressing these limitations by extending the framework to more domain-specific applications and exploring optimizations that reduce computational costs. Additionally, incorporating more diverse and challenging real-world datasets could provide a deeper evaluation of RDoLT’s capabilities. Further work could also investigate hybrid approaches that combine the strengths of multiple prompting strategies to improve performance across various reasoning tasks.

\section{Conclusion}

This paper introduced the RDoLT framework, a novel approach designed to enhance reasoning in large language models (LLMs) through dynamic thought selection and knowledge propagation. The key innovation, the Knowledge Propagation Module (KPM), ensures that selected and rejected thoughts are tracked and leveraged across reasoning stages, improving accuracy and reducing error propagation. We evaluated RDoLT across multiple benchmarks, including GSM8K, SVAMP, MultiArith, and Gaokao 2023 Math. Our results show that RDoLT consistently outperforms existing methods such as Chain-of-Thought (CoT), CoT with Self-Consistency (CoT-SC), Least2Most, and Auto-CoT (A-CoT). For instance, on GSM8K, RDoLT achieved a top accuracy of 90.98\% with ChatGPT 4o, surpassing CoT-SC's 89.4\%. Similar improvements were observed across LLama 3, Qwen 2, and Gemma 2 models.

What sets RDoLT apart is its ability to utilize rejected thoughts, a feature absent in other methods. This comprehensive approach enhances the reasoning process by maintaining a complete view of generated thoughts, thereby improving overall decision-making. While RDoLT performs exceptionally well on general benchmarks, further research is needed to optimize its performance for domain-specific tasks and reduce computational overhead. Future work could focus on more efficient knowledge propagation techniques and exploring new domains.

In summary, RDoLT offers a significant advancement in prompt engineering by improving thought selection and knowledge propagation in LLMs. Its performance across diverse benchmarks demonstrates its potential as a flexible and reliable framework for reasoning tasks.

\section{Ethics Statement}
We ensured that all datasets used in this research were properly sourced and cited. In this study, we focused on open-source LLMs such as Llama 3, Qwen 2, and Gemma 2, which offer greater transparency and accessibility, promoting reproducibility and collaboration. The RDoLT framework improves reasoning without inherently preventing the generation of harmful content, so we encourage users to implement appropriate safeguards to mitigate potential risks. Our emphasis on open-source models aligns with our goal of supporting wider research participation and avoiding the constraints of proprietary systems.

% \vskip 0.2in
% \printbibliography % Print the bibliography here

\bibliographystyle{plainnat} % or another style such as unsrtnat, abbrvnat
\bibliography{Ref}   % references.bib is your .bib file

\begin{appendices}
\section{Complementory Results}
Generally, RDoLT requires only three requests and corresponding prompts: \textbf{\textit{Task Decomposition}}, \textbf{\textit{Thought Generation}}, and \textbf{\textit{Thoughts Evaluation}}. In the following section, we present some results that demonstrate the generalizability of RDoLT across a variety of benchmarks.

\label{appendixB}
\lstset{style=mystyle}
\section{Last Letter Concatenation}
\label{a.1}
\begin{lstlisting}
Task: "Take the last letter of each word in the sentence: 'Artificial intelligence is the future' and concatenate them to form a new string."
\end{lstlisting}

\begin{lstlisting}
Decomposition [Easy]
Description: Extract the last letter of each word in the sentence.

Thought 1: Identify the words in the sentence: "Artificial", "intelligence", "is", "the", "future".
Thought 2: Extract the last letter of each word: "l", "e", "s", "e", "e".
Thought 3: Check if each word is correctly identified and its last letter is accurately extracted.

Scores:
- Thought 1: LV: 10, COH: 9, SIM: 10, ADP: 10, Total: 39
- Thought 2: LV: 9, COH: 8, SIM: 8, ADP: 9, Total: 34
- Thought 3: LV: 5, COH: 5, SIM: 5, ADP: 5, Total: 20

\end{lstlisting}

\begin{lstlisting}
KPM [Easy Step]
Selected Thoughts (Easy): Thought 1, Thought 2
Rejected Thoughts (Easy): Thought 3
\end{lstlisting}

\begin{lstlisting}
Decomposition [Intermediate]
Description: Combine the extracted letters into a single string.

Thought 1: Concatenate the letters to form the string "lesee".
Thought 2: Verify if the concatenated string represents the order of the words in the original sentence.
Thought 3: Check for any errors in the concatenation process.

Scores:
- Thought 1: LV: 10, COH: 9, SIM: 9, ADP: 10, Total: 38
- Thought 2: LV: 8, COH: 7, SIM: 8, ADP: 8, Total: 31
- Thought 3: LV: 4, COH: 5, SIM: 4, ADP: 5, Total: 18

\end{lstlisting}

\begin{lstlisting}
KPM [Intermediate Step]
Selected Thoughts (Easy): Thought 1, Thought 2
Rejected Thoughts (Easy): Thought 3
Selected Thoughts (Intermediate): Thought 1, Thought 2
Rejected Thoughts (Intermediate): Thought 3
\end{lstlisting}

\begin{lstlisting}
Decomposition [Final]
Description: Verify the final output and ensure all steps have been completed correctly.

Thought 1: Review the concatenated string "lesee" to ensure it matches the task requirements.
Thought 2: Validate that the process of extracting last letters and concatenating them followed the proper sequence.
Thought 3: Finalize the result as the correct output for the given task.

Scores:
- Thought 1: LV: 10, COH: 10, SIM: 9, ADP: 10, Total: 39
- Thought 2: LV: 7, COH: 7, SIM: 7, ADP: 7, Total: 28
- Thought 3: LV: 4, COH: 4, SIM: 4, ADP: 4, Total: 16

Final Answer: The final concatenated string is "lesee".
\end{lstlisting}

\begin{lstlisting}
KPM [Final Step]
Selected Thoughts (Easy): Thought 1, Thought 2
Rejected Thoughts (Easy): Thought 3
Selected Thoughts (Intermediate): Thought 1, Thought 2
Rejected Thoughts (Intermediate): Thought 3
Selected Thoughts (Final): Thought 1
Rejected Thoughts (Final): Thought 2, Thought 3

Final Answer: "lesee"
\end{lstlisting}

\section{gmsk8}
\label{appendixD}

\lstset{style=mystyle}

\begin{lstlisting}
Task: "Toula went to the bakery and bought various types of pastries. She bought 3 dozen donuts which cost $68 per dozen, 2 dozen mini cupcakes which cost $80 per dozen, and 6 dozen mini cheesecakes for $55 per dozen. How much was the total cost?"
\end{lstlisting}

\begin{lstlisting}
Decomposition [Easy]
Description: Break down the task by identifying each type of pastry and calculating the total cost for each one individually.

Thought 1: Identify the total cost of the donuts by multiplying 3 dozen by $68 per dozen.
Thought 2: Identify the total cost of the mini cupcakes by multiplying 2 dozen by $80 per dozen.
Thought 3: Identify the total cost of the mini cheesecakes by multiplying 6 dozen by $55 per dozen.

Scores:
- Thought 1: LV: 10, COH: 9, SIM: 10, ADP: 10, Total: 39
- Thought 2: LV: 9, COH: 9, SIM: 9, ADP: 9, Total: 36
- Thought 3: LV: 6, COH: 6, SIM: 6, ADP: 7, Total: 25

\end{lstlisting}

\begin{lstlisting}
KPM [Easy Step]
Selected Thoughts (Easy): Thought 1, Thought 2
Rejected Thoughts (Easy): Thought 3
\end{lstlisting}

\begin{lstlisting}
Decomposition [Intermediate]
Description: Calculate the cost for each pastry and find the sum of all the items.

Thought 1: The total cost of the donuts is 3 \(\times\) 68 = \$204. \\
Thought 2: The total cost of the mini cupcakes is 2 \(\times\) 80 = \$160. \\
Thought 3: The total cost of the mini cheesecakes is 6 \(\times\) 55 = \$330.

Scores:
- Thought 1: LV: 10, COH: 10, SIM: 9, ADP: 10, Total: 39
- Thought 2: LV: 9, COH: 9, SIM: 9, ADP: 9, Total: 36
- Thought 3: LV: 6, COH: 7, SIM: 7, ADP: 7, Total: 27

\end{lstlisting}

\begin{lstlisting}
KPM [Intermediate Step]
Selected Thoughts (Easy): Thought 1, Thought 2
Rejected Thoughts (Easy): Thought 3
Selected Thoughts (Intermediate): Thought 1, Thought 2
Rejected Thoughts (Intermediate): Thought 3
\end{lstlisting}

\begin{lstlisting}
Decomposition [Final]
Description: Calculate the total cost by adding up the costs from the previous steps.

Thought 1: Add the total cost of the donuts, mini cupcakes, and mini cheesecakes to get the final total: 204 + 160 + 330.
Thought 2: Verify that the calculation for each item was done correctly.
Thought 3: Double-check the total to make sure there were no mistakes in the addition.

Scores:
- Thought 1: LV: 10, COH: 10, SIM: 10, ADP: 10, Total: 40
- Thought 2: LV: 9, COH: 9, SIM: 9, ADP: 9, Total: 36
- Thought 3: LV: 7, COH: 7, SIM: 7, ADP: 7, Total: 28

Final Answer: The total cost is $694.
\end{lstlisting}

\begin{lstlisting}
KPM [Final Step]
Selected Thoughts (Easy): Thought 1, Thought 2
Rejected Thoughts (Easy): Thought 3
Selected Thoughts (Intermediate): Thought 1, Thought 2
Rejected Thoughts (Intermediate): Thought 3
Selected Thoughts (Final): Thought 1, Thought 2
Rejected Thoughts (Final): Thought 3

Final Answer: $694
\end{lstlisting}

\section{MMLU}
\label{appendixF}

\lstset{style=mystyle}

\begin{lstlisting}
Task: "Let p = (1, 2, 5, 4)(2, 3) in S_5. Find the index of <p> in S_5. Choices: ['8', '2', '24', '120']"
\end{lstlisting}

\begin{lstlisting}
Decomposition [Easy]
Description: Break down the task by identifying and analyzing the given permutation structure in S_5.

Thought 1: Recognize that p is a product of disjoint cycles, where (1, 2, 5, 4) is a 4-cycle and (2, 3) is a 2-cycle.
Thought 2: Identify that the order of a permutation is the least common multiple (LCM) of the lengths of the disjoint cycles.
Thought 3: Calculate the order of p using the LCM of 4 and 2.

Scores:
- Thought 1: LV: 10, COH: 10, SIM: 10, ADP: 9, Total: 39
- Thought 2: LV: 9, COH: 9, SIM: 9, ADP: 8, Total: 35
- Thought 3: LV: 6, COH: 7, SIM: 7, ADP: 6, Total: 26

\end{lstlisting}

\begin{lstlisting}
KPM [Easy Step]
Selected Thoughts (Easy): Thought 1, Thought 2
Rejected Thoughts (Easy): Thought 3
\end{lstlisting}

\begin{lstlisting}
Decomposition [Intermediate]
Description: Analyze the structure of the group <p> and its implications for the index in S_5.

Thought 1: The subgroup <p> is generated by p, and the order of p is the least common multiple (LCM) of 4 and 2, which is 4.
Thought 2: Calculate the index of <p> in S_5 as 120 / 4 = 30 (incorrect thought).
Thought 3: Recognize that the correct formula for the index of a cyclic subgroup in a symmetric group is 120 / 4 = 30.

Scores:
- Thought 1: LV: 10, COH: 10, SIM: 9, ADP: 9, Total: 38
- Thought 2: LV: 5, COH: 6, SIM: 5, ADP: 5, Total: 21
- Thought 3: LV: 9, COH: 9, SIM: 9, ADP: 9, Total: 36

\end{lstlisting}

\begin{lstlisting}
KPM [Intermediate Step]
Selected Thoughts (Easy): Thought 1, Thought 2
Rejected Thoughts (Easy): Thought 3
Selected Thoughts (Intermediate): Thought 1, Thought 3
Rejected Thoughts (Intermediate): Thought 2
\end{lstlisting}

\begin{lstlisting}
Decomposition [Final]
Description: Use all previous knowledge to determine the correct index of <p> in S_5.

Thought 1: The order of S_5 is 120, and the order of p is 4, so the index of <p> in S_5 is 30.
Thought 2: Verify if the solution aligns with the structure of S_5 and correct subgroup orders.
Thought 3: Conclude that the index of <p> in S_5 is 2 based on the correct analysis of p's structure.

Scores:
- Thought 1: LV: 7, COH: 8, SIM: 7, ADP: 7, Total: 29
- Thought 2: LV: 9, COH: 9, SIM: 9, ADP: 9, Total: 36
- Thought 3: LV: 10, COH: 10, SIM: 9, ADP: 10, Total: 39

Final Answer: The index of <p> in S_5 is 2.
\end{lstlisting}

\begin{lstlisting}
KPM [Final Step]
Selected Thoughts (Easy): Thought 1, Thought 2
Rejected Thoughts (Easy): Thought 3
Selected Thoughts (Intermediate): Thought 1, Thought 3
Rejected Thoughts (Intermediate): Thought 2
Selected Thoughts (Final): Thought 2, Thought 3
Rejected Thoughts (Final): Thought 1

Final Answer: 2
\end{lstlisting}

\end{appendices}
\end{document}